# The Enemy from Within: A Study of Political Delegitimization Discourse in Israeli Political Speech

**Naama Rivlin-Angert*** **and** **Guy Mor-Lan***
The Hebrew University of Jerusalem
{naama.rivlin-angert|guy.mor}@mail.huji.ac.il

## Abstract

We present the first large-scale computational study of political delegitimization discourse (PDD), defined as symbolic attacks on the normative validity of political entities. We curate and manually annotate a novel Hebrew-language corpus of 10,410 sentences drawn from parliamentary speeches (1993-2023), Facebook posts, and leading news outlets (2018-2021), of which 1,812 instances (17.4%) exhibit PDD and 642 carry additional annotations for intensity, incivility, target type, and affective framing. We introduce a two-stage classification pipeline, and benchmark finetuned encoder models and decoder LLMs. Our best model (DictaLM 2.0) attains an $F_1$ of 0.74 for binary PDD detection and a macro-$F_1$ of 0.67 for classification of delegitimization characteristics. Applying this classifier to longitudinal and cross-platform data, we see a marked rise in PDD over three decades, higher prevalence on social media versus parliamentary debate, greater use by male politicians than by their female counterparts, and stronger tendencies among right-leaning actors, with pronounced spikes during election campaigns and major political events. Our findings demonstrate the feasibility and value of automated PDD analysis for analyzing democratic discourse.[1]

## 1 Introduction

Legitimacy is a fundamental pillar of democratic governance, underpinning the consent and participation of citizens in political processes (Suchman, 1995; Weber, 1958). In recent decades, however, democratic norms have eroded amid rising affective polarization and the growing prevalence of hostile rhetoric (Iyengar and Krupenkin, 2018; Levitsky and Ziblatt, 2019). Delegitimization discourse, the strategic portrayal of political opponents as unworthy of normative inclusion, has emerged as a powerful tool in political competition, yet it remains underexplored in computational social science. While adjacent research streams have advanced the automated detection of incivility, hate speech, and toxicity (Frimer et al., 2023; Jahan and Oussalah, 2023), these frameworks typically target surface-level norm violations rather than the symbolic function of undermining political legitimacy itself.

In this paper, we propose Political Delegitimization Discourse (PDD) as a distinct analytical category and introduce a computational pipeline for its large-scale detection and characterization. We curate and annotate a diverse Hebrew-language corpus of over 10,000 sentences drawn from parliamentary speeches, social media, and news media, capturing the multifaceted nature of PDD across institutional and digital communication channels. Leveraging state-of-the-art encoder and decoder models, our two-stage classifier first identifies instances of PDD and then predicts their intensity, rhetorical attributes, and target types. Applying this framework to longitudinal and cross-platform data, we uncover temporal trends, platform-specific variations, and demographic patterns that shed new light on the dynamics of delegitimization in Israeli political discourse.

### 1.1 Legitimization and Delegitimization

Legitimacy and justification are fundamental prerequisites for any entity that aspires to govern, exercise force, and impose obligations on its constituents. Legitimacy is not a static condition, but rather it continuously evolves through social processes. Legitimization occurs when previously contested actions, policies, or claims undergo reclassification and become accepted, while *delegitimization* refers to the opposite process, where what were once legitimate are stripped of their normative validity and are framed as unacceptable

* Both authors contributed equally to this work.
[1] https://github.com/guymorlan/pdd/

(Kelman, 2001).

Discursive (de)legitimization refers to social and political actions carried out through written or spoken language (van Dijk, 1998). This perspective treats (de)legitimization as a communicative action, reflecting dynamic and ongoing negotiations over legitimacy (Cap, 2008; Chilton, 2004; van Dijk, 1997; van Leeuwen, 2007; Reyes, 2011). These processes do not emerge independently; they are triggered by actions or declarations from authority figures operating within political, religious, judicial, or other institutions (Suchman, 1995).

While previous social science research has investigated discursive delegitimization, it has primarily centered on intergroup dynamics and marginalization of social groups (Bar-Tal, 1989, 1990; Baryla et al., 2015; Holland and Wright, 2017; Tileagă, 2007; Rinnawi, 2007; Volpato et al., 2010; Winter, 2016). However, since political identities fundamentally function as a form of social identity (Green et al., 2004; Mason, 2018), recent work has extended this line of inquiry to the delegitimization of political identities (Rivlin-Angert, 2025).

## 2 Analyzing PDD

Political Delegitimization Discourse (PDD) is defined as discourse that seeks to undermine the legitimacy of political identities through attacks against their *symbolic aspects*, rather than criticizing specific policies or actions. The objective of PDD is to establish political dominance and suppress alternative narratives by narrowing the boundaries of acceptable discourse and excluding specific ideas and ideologies from being considered legitimate (Rivlin-Angert, 2025).

To date, however, existing research on delegitimizing statements within political discourse has relied primarily on manual coding of small datasets, lacking scalability and longitudinal insight (Berrocal, 2019; Baldi and Franco, 2015; Baldi et al., 2019; Egelhofer et al., 2021; Gadavanij, 2020; Ross and Rivers, 2017; Screti, 2013). To the best of our knowledge, no computational framework has yet been developed to identify and track PDD systematically over time or across political actors. This paper addresses that gap by introducing a computational pipeline for detecting PDD in a large-scale corpus of Israeli political discourse.

### 2.1 Research Questions

Our analysis focuses on five research questions. First, prior work has documented a steady rise in affective polarization (Iyengar and Krupenkin, 2018; Iyengar et al., 2019; Gidron et al., 2020) and a decline in democratic norms such as civility and mutual respect (Levitsky and Ziblatt, 2019; Svolik, 2019). These trends suggest a shift toward more hostile political rhetoric. Therefore, we ask:

**RQ1. Temporal trends.** *How has PDD prevalence evolved over time?*

Social media enables politicians to bypass traditional gatekeepers like journalists and editors, offering direct and unfiltered access to the public (Bartlett, 2014; Ekman and Widholm, 2015; Parmelee and Bichard, 2011). Indeed, studies on incivility and hate speech point to a rise in norm-violating rhetoric, particularly in online platforms (Theocharis et al., 2020). Given this unmediated communication environment, we ask:

**RQ2. Platform differences.** *How does PDD vary between social media platforms and parliamentary speech?*

Prior research highlights systematic gender differences in political communication, with female politicians often adopting more moderate or conciliatory styles compared to their male counterparts (Karpowitz and Mendelberg, 2014; Haselmayer et al., 2022). We therefore ask:

**RQ3. Gender differences.** *Are there gender-based differences in PDD frequency or intensity?*

Previous research has shown that incivility and polarizing discourse vary across ideological lines, with right-leaning actors often using more norm-violating rhetoric (Skytte, 2021; van Elsas and Fiselier, 2023). We therefore ask:

**RQ4. Bloc differences.** *How does the prevalence of PDD differ across political blocs?*

The use of PDD often depends on political context (Rivlin-Angert, 2025). During high-stakes moments, such as election campaigns, politicians face heightened incentives to delegitimize opponents in order to mobilize supporters, shape public opinion, or weaken rivals. In contrast, post-election periods typically shift focus to coalition-building and negotiation, where such rhetoric may undermine future alliances and become less advantageous. This leads us to the following question:

**RQ5. Contextual variation.** *How does PDD shift around elections and major political events?*

## 3 Related Work

The automatic detection of PDD lies at the intersection of several research streams in computational social science and natural language processing. While no prior work directly models PDD as a standalone task, adjacent research has focused on computational modeling of incivility, intolerance, hate speech, and toxicity. In addition, tasks such as stance detection and targeted sentiment analysis have sought to model how speakers evaluate or position themselves toward specific entities. This section reviews key contributions across these areas and highlights the conceptual differences between them and PDD.

*Incivility* is generally characterized as rude, disrespectful, or norm-violating language (e.g., name-calling, profanity, hyperbole) directed at individuals or groups (Rossini, 2022; Frimer et al., 2023). Early approaches used supervised classifiers to identify abusive language directed at politicians (e.g., Theocharis et al., 2020; Rheault et al., 2019; Da San Martino et al., 2020). Recent efforts have improved granularity; for example, Frimer et al. (2023) demonstrate how language models can uncover civility gradients in ideological exchanges.

*Hate speech*, on the other hand, is typically defined as explicitly abusive content, animosity, or disparagement of an individual or a group on account of a group characteristic (such as race, color, national origin, sex, disability, religion, or sexual orientation) (Nockleby, 1994). Progress in NLP has spurred extensive research efforts focused on automating the detection of hate speech in textual data (Jahan and Oussalah, 2023; Saleh et al., 2023).

*Toxicity* is used as a broader umbrella term (Calvo et al., 2023; Hansen, 2023; Gervais et al., 2025) that overlaps with both incivility and hate speech in capturing generally harmful or abusive language (Buell, 1998). Related strands of work on dehumanization (Mendelsohn et al., 2020; Burovova and Romanyshyn, 2024), populist rhetoric (Klamm et al., 2023; Erhard et al., 2025; Tao et al., 2024; Zhou et al., 2024), and ad hominem attacks (Delobelle et al., 2019; Habernal et al., 2018) have modeled the ways language constructs in-groups and out-groups.

PDD differs fundamentally from incivility, hate speech and toxicity: it is defined not by the presence of profanity or insults but by its symbolic goal of denying an actor's right to political inclusion; it is not solely about the *how* something is said (e.g., whether it is rude or offensive), but rather about *what* it does: denies the political legitimacy of a person, group, or institution.

Beyond these categories, NLP has developed adjacent tasks such as stance detection and targeted sentiment analysis. Stance detection aims to classify whether a text expresses support, opposition, or neutrality toward a given proposition or entity (ALDayel and Magdy, 2021; Mohammad et al., 2017). Targeted sentiment analysis similarly seeks to identify the polarity of sentiment directed toward specific targets (Zhang et al., 2016).

PDD also differs from stance detection and targeted sentiment analysis. While stance detection and targeted sentiment analysis provide valuable tools for measuring attitudes, they capture a general evaluative direction and not the specific symbolic act of denying political legitimacy. For example, a politician may express negative sentiment or opposition toward another, without necessarily questioning their right to participate in the political system.

Although PDD can manifest through aggressive language, it may operates through subtler means - such as strategic framing, implication, or selective factual emphasis - that question a person's or group's legitimacy without violating surface-level norms. Because these instances lack overt hostility, previous classifiers frequently overlook them, underestimating the true prevalence of delegitimizing rhetoric. To address this gap, we offer a precise operational definition of PDD grounded in political theory, and introduce a dedicated two-stage detection pipeline, trained on a longitudinal Hebrew corpus. By modeling PDD as a distinct NLP task, we expand the community's ability to detect subtle forms of harmful political speech and provide a new computational lens for studying democratic erosion and affective polarization at scale.

## 4 Annotation Scheme

We developed an annotation scheme grounded in a conceptual definition of PDD as discourse aimed at discrediting political groups or actors by attacking their symbolic and affective dimensions. Table 1 presents the attributes of our sentence-level annotation scheme.

The annotation process was designed in two stages. In the first stage, individual sentences are annotated for the binary presence or absence of political delegitimization discourse (PDD). A sentence is considered positive for PDD if it targets a political actor, group, or institution and conveys a hostile characterization not grounded in policy critique but in symbol-based delegitimization. Typical indicators include expressions of disgust or ridicule (such as pejorative nicknames), claims that the target poses a threat to the state or society, denial of the target's right to political participation, or comparisons to stigmatized out-groups (e.g., Nazis, terrorists).[2]

| **1. Political Delegitimization Discourse** | |
|---|---|
| Delegit. (T/F) | Sentence expresses PDD. |
| **2. Attributes of PDD (if Delegit.=T)** | |
| Intensity (0-2) | Strength of PDD (2 = strongest). |
| Incivility (T/F) | Mockery, swearing, insults. |
| Outgroup (T/F) | Casts target as external "enemy" (e.g., "the Fascists"). |
| Common good (T/F) | Invokes threat to society at large. |
| **3. Target Attributes (if Delegit.=T)** | |
| Group (T/F) | Target is a social or political group (e.g., "Leftists"). |
| Person (T/F) | Target is an individual (e.g., "Benjamin Netanyahu"). |
| Institute (T/F) | Target is an organization (e.g., "Supreme Court"). |
| Target spans | Token spans marking the delegitimized referent. |

Table 1: Overview of Annotation Scheme

In the second stage, sentences identified as PDD-positive are further annotated along several dimensions to capture variation in rhetorical form and intensity. This includes a rating of delegitimization intensity on a three-point scale (0 = weak, 2 = strong). Additional binary annotations recorded whether the sentence includes incivility features (such as mockery, slander, or profanity); whether it associates the target with an illegitimate or stigmatized out-group; and whether it frames the target as harmful to the common good. Finally, the annotation also records attributes of the delegitimized target, including binary labels for whether the target is a person, group, or an institution. Additionally, the annotation identifying all spans in the sentence that reference the target if it is explicitly referenced.

We illustrate PDD with several examples from our data:

1. *"I call to investigate Nitzan Horowitz on suspicion of betrayal against the State of Israel."*

This instance targets a political actor and questions their symbolic alignment with the nation.

2. *"The truth must be told, they may be part of us, but they never stop destroying us from the inside."*

This statement targets a broad political group and portrays them as internal saboteurs, suggesting that they are fundamentally disloyal and harmful, and constructing them as an internal threat.

3. *"To satisfy the hatred of their supporters and to fuel the false theory of the Mizrahi people, Barashi and Distel are willing to trample those who survived the Holocaust."*

This instance attributes malicious intent to the actions of political actors, suggesting they are driven by hatred and willing to violate sacred moral boundaries.

Importantly, critique or negative attitudes towards policy issues is not in itself a sufficient condition for PDD, even if phrased in harsh terms. For example, the sentence *"This [budget] is an example of negligence and promiscuous conduct, endangering both the economy and society in Israel."* is not annotated as PDD.

## 5 Dataset Construction

Our dataset contains sentences sampled from three sources:

**Facebook posts.** We compile a corpus of official Facebook posts, including all verified accounts of Israeli political parties, incumbent members of the Israeli parliament (MKs), and additional viable candidates.[3] Data was collected via the CrowdTangle API between December 2018 and April 2021. This corpus contains 57K posts by 206 political accounts, comprising 322K sentences.

**Parliamentary speeches.** We utilize the IsraParlTweet dataset (Mor-Lan et al., 2024), a full archive of Knesset floor speeches from 1992-2023.

**News data.** We collected news articles on a daily cadence from 38 leading Hebrew-language outlets spanning Israel's political spectrum. These include all major digital news websites (e.g., Ynet, Walla,

[2] The full annotation guidelines including detailed criteria are provided in Appendix A.

[3] We define viable candidates as those who were next in line after the elected MKs. Specifically, the next 15% of each party's candidate list following those who won seats in the most recent election.

Mako) as well as prominent sectoral and niche outlets (e.g., Channel 7, Srugim). Articles were collected via the Webz.io API between October 2018 and November 2022, resulting in 1.7M articles.

Delegitimization was annotated as a binary category (T/F) by three human annotators, including the lead author and two graduate students. A sample of 170 sentences was used for inter-coder reliability, showing a high average correlation of 0.91 between the annotators and an average Cohen's Kappa of 0.82, indicating substantial agreement. Disagreements were decided by the lead author in joint consultation. In the second stage, a subset of sentences containing PDD from Facebook posts and news media was additionally annotated by the lead author for delegitimization characteristics and span-level indicators of the target of delegitimization.

This process results in a novel annotated corpus of 10,410 Hebrew-language sentences, of which 1,812 instances (17.4%) exhibit PDD and 642 sentences carry additional annotations for intensity, incivility, target type, and affective framing. The resulting dataset was split into a train set (70%), validation set (15%) and test set (15%).

### 5.1 Descriptive Statistics

Table 2 shows annotated sentences by source, and the proportions and counts of PDD features are given in Table 3.

| Source | Count | Percentage |
|---|---|---|
| Facebook | 6690 | 64.27% |
| Knesset | 2504 | 24.06% |
| News media | 1216 | 11.68% |
| Total | 10410 | 100% |

Table 2: Breakdown of Items by Source

## 6 Modeling and Experiments

We examine a set of encoder models and LLM decoders in the 2B-9B parameters range. For encoders, we use the multilingual mBERT (Devlin et al., 2019), and Hebrew-targeted encoders AlephBERT (Seker et al., 2022), HeRO (Shalumov and Haskey, 2023) and the base and large variants of DictaBERT (Shmidman et al., 2023). All encoder models are trained for up to 10 epochs with three learning rates (1e-5, 3e-5, 5e-5), choosing the best performing checkpoint on the validation set. For decoder LLMs, we use Gemma 2 2B and 9B (Gemma Team, 2024), Qwen3 8B (Qwen Team, 2025), and Hebrew-targeted DictaLM2.0 (Shmidman et al., 2024). All decoders are fine-tuned with QLORA (Dettmers et al., 2023) for up to 6 epochs with two learning rates (1e-5, 1e-4). For prompt templates, see Appendix B.

| **Feature** | **Count** | **Percent** |
|---|---|---|
| Delegitimization | 1812 | 17.4% |
| **A subset where Delegit.=True, N=642** | | |
| Incivility | 157 | 25.0% |
| Common good | 147 | 23.4% |
| Outgroup | 147 | 23.4% |
| Target: Group | 174 | 27.7% |
| Target: Person | 271 | 43.2% |
| Target: Institute | 163 | 26.0% |
| **Feature** | **Mean** | **SD.** |
| Intensity | 1.2 | 0.73 |
| **Feature** | **Total spans** | **% with target span** |
| Target spans | 471 | 54.9% |

Table 3: Descriptive Statistics

We model PDD as a two-stage pipeline:

1. **Binary PDD classifier**. Encoders are trained with three types of loss: default, class weights and focal loss (Lin et al., 2017), and decoders are trained to produce a Hebrew True/False label.

2. $2^{nd}$ stage tasks (only where PDD=1).

   (a) **Multi-task classifier for properties of PDD.** Encoder models are trained with multi-task loss (6 binary labels and one multi-class label with three categories). Decoders are fine-tuned to produce a JSON representation of the outputs with Hebrew language keys and numeric values.

   (b) **Span classifier for target of PDD**. We utilize the decoder models only, fine-tuned to reproduce the input sentence with %%% tokens wrapping any mentions of PDD targets.

Figure 1 shows the two-stage pipeline for PDD detection, including both identification and classification stages. The results for the first-stage task, the properties of delegitimization task, and the target span identification task are presented in tables

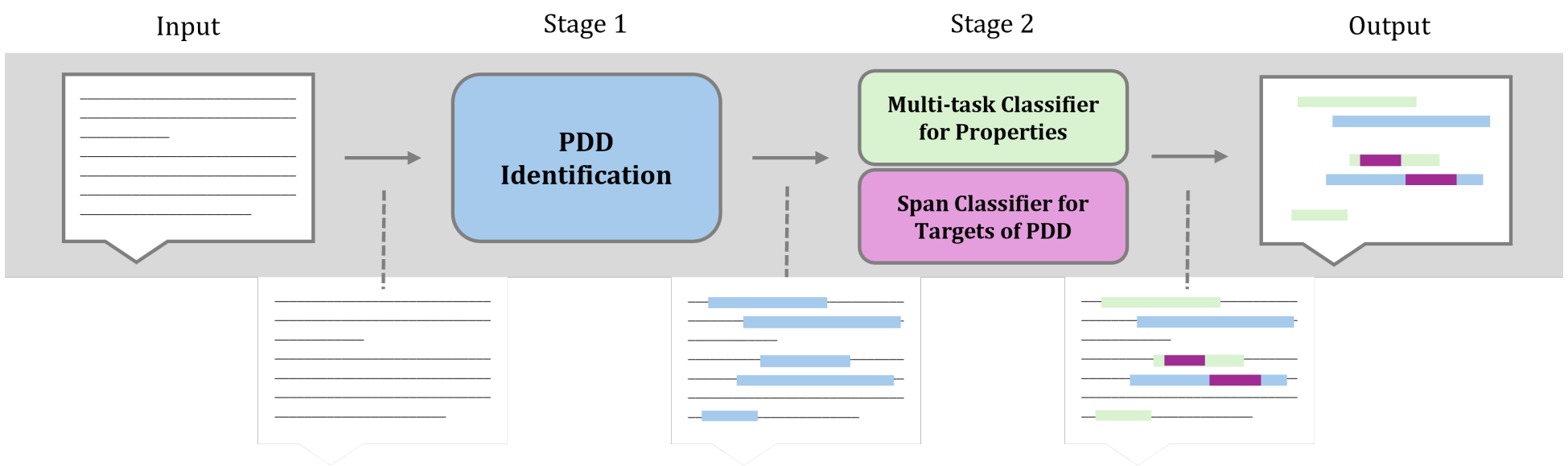

Figure 1: Two-stage Pipeline for PDD Detection, Including Identification and Classification

4, 5, 6. For the first-stage, the DictaLM2.0 decoder is the most performant with an $F_1$ score of 0.74, followed by the DictaBERT-base encoder with an $F_1$ score of 0.71. DictaLM2.0 is also the most performant in the delegitimization characteristics task, achieving an average $F_1$ score of 0.67 (although only achieving best results on 2/7 labels) and on the target mentions task, achieving a span-level $F_1$ score of 0.67. DictaLM2.0 checkpoints are thus utilized in the following analysis.

| Model | Loss | LR | Acc. | P | R | $F_1$ |
|---|---|---|---|---|---|---|
| **Decoder LLMs** | | | | | | |
| DictaLM2.0 | default | 1e-05 | **0.905** | **0.756** | 0.714 | **0.735** |
| Gemma-2-9B | default | 1e-05 | 0.896 | 0.746 | 0.655 | 0.698 |
| Gemma-2-2B | default | 1e-05 | 0.878 | 0.705 | 0.582 | 0.637 |
| Qwen-3-8B | default | 1e-05 | 0.780 | 0.367 | 0.268 | 0.310 |
| **Encoders** | | | | | | |
| HeRo | default | 1e-05 | 0.887 | 0.728 | 0.617 | 0.668 |
| DictaBERT-B | default | 1e-05 | 0.889 | 0.676 | **0.756** | 0.714 |
| DictaBERT-L | class w. | 3e-05 | 0.892 | 0.723 | 0.666 | 0.693 |
| mBERT | default | 1e-05 | 0.859 | 0.635 | 0.551 | 0.590 |
| AlephBERT | default | 1e-05 | 0.887 | 0.678 | 0.735 | 0.706 |

Table 4: $1^{st}$ Stage Results

## 7 Analysis

To examine the research questions, we use our full Facebook dataset and the full parliamentary portion of IsraParlTweet. We classify all sentences with the fine-tuned DictaLM2.0 checkpoints and compute shares of PDD and of PDD characteristics. For all analyses, the proportions were computed by first aggregating at the speaker level to avoid over-weighting more active politicians.

To evaluate changes in PDD over time (RQ1), we analyze 30 years of parliamentary speeches (1993-2023), computing the mean share of delegitimizing sentences aggregated by half-year periods (see Figure 2). The results reveal a marked increase in PDD over the period. While the 1990s and early 2000s show relatively stable and moderate levels, a decline appears around 2008, followed by a gradual increase throughout the 2010s. Beginning in 2020, the prevalence of PDD rose sharply, reaching the highest levels in 2023. We find that the mean share of PDD increased from 6.6% in 1992–2019 to 9.8% in 2019–2023. This difference in means is statistically significant ($t = -105.24$, $p < 0.01$, $|d| = 0.13$). These patterns suggest a long-term upward trend in delegitimizing rhetoric within Israeli parliamentary discourse.

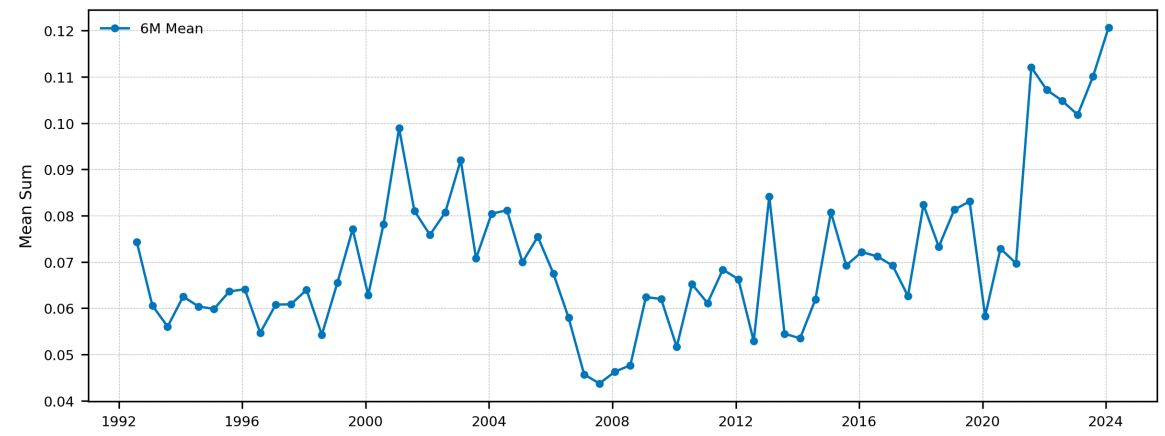

Figure 2: PDD Share in Knesset Speeches

To address RQ2, we compare the prevalence and characteristics of PDD across two communication arenas: parliamentary speeches and social media posts, during the overlapping period of 2018-2021. Within each platform, we compute the mean share of PDD sentences and of each PDD characteristic. Results are presented in Table 7. While the overall prevalence of PDD is similar across platforms (7.33% on Facebook vs. 7.05% in the Knesset), the characteristics of the discourse diverge notably. PDD posts on Facebook exhibit consistently higher rates of incivility, references to outgroups, accusations of harm to the common good, and more frequent targeting of individuals, groups, and institutions. The average intensity of PDD is also higher on Facebook (1.23) than in Knesset speeches (1.00), indicating that online PDD tends to be not only more rhetorically aggressive but also more symbolically loaded.

To examine gender-based differences in the use of PDD (RQ3), we compare the distribution of mean PDD scores aggregated by speaker gender in

| Model | LR | Intensity $F_1$ | Incivility $F_1$ | Group $F_1$ | Person $F_1$ | Outgroup $F_1$ | Common-good $F_1$ | Institute $F_1$ | Avg. $F_1$ |
|---|---|---|---|---|---|---|---|---|---|
| **Decoder LLMs** | | | | | | | | | |
| DictaLM2.0 | 1e-05 | 0.587 | **0.528** | **0.776** | 0.769 | 0.737 | 0.717 | 0.533 | **0.664** |
| Qwen 3 8B | 1e-05 | 0.458 | 0.245 | 0.567 | 0.624 | 0.588 | 0.444 | 0.356 | 0.469 |
| Gemma 2 2B | 1e-05 | 0.522 | 0.377 | 0.677 | 0.731 | 0.679 | 0.642 | 0.625 | 0.608 |
| Gemma 2 9B | 1e-05 | 0.583 | 0.275 | 0.649 | **0.792** | 0.792 | **0.720** | 0.622 | 0.633 |
| **Encoders** | | | | | | | | | |
| HeRo | 3e-05 | 0.448 | 0.400 | 0.704 | 0.755 | **0.800** | 0.465 | 0.566 | 0.591 |
| DictaBERT-B | 5e-05 | **0.596** | 0.360 | 0.687 | 0.758 | 0.733 | 0.630 | **0.694** | 0.637 |
| DictaBERT-L | 5e-05 | 0.439 | 0.364 | 0.697 | 0.777 | 0.690 | 0.615 | 0.667 | 0.607 |
| AlephBERT | 5e-05 | 0.518 | 0.327 | 0.667 | 0.745 | 0.679 | 0.390 | 0.488 | 0.545 |
| mBERT | 3e-05 | 0.398 | 0.178 | 0.627 | 0.653 | 0.632 | 0.419 | 0.510 | 0.488 |

Table 5: 2nd Stage Results: Delegitimization Characteristics

| Model | LR | Precision | Recall | $F_1$ | #TP |
|---|---|---|---|---|---|
| DictaLM2.0 | 1e-05 | **0.750** | **0.600** | **0.667** | **51** |
| Gemma 2 2B | 1e-05 | 0.685 | 0.435 | 0.532 | 37 |
| Gemma 2 9B | 1e-05 | 0.639 | 0.541 | 0.586 | 46 |
| Qwen 3 8B | 1e-05 | 0.515 | 0.400 | 0.450 | 34 |

Table 6: 2nd Stage Results: Span Target Mentions

| **Metric** | **Knesset** | **Facebook** | **Sig.** | $\|d\|$ |
|---|---|---|---|---|
| PDD | 7.05% | 7.33% | *** | 0.05 |
| **Characteristics of PDD (when PDD = 1)** | | | | |
| Incivility | 17.3% | 19.1% | *** | 0.12 |
| Outgroup | 7.9% | 15.4% | *** | 0.19 |
| Common Good | 8.5% | 15.6% | *** | 0.16 |
| Target: Group | 15.7% | 18.5% | *** | 0.13 |
| Target: Person | 36.9% | 41.5% | *** | 0.09 |
| Target: Institute | 17.0% | 20.8% | *** | 0.08 |
| Intensity (avg.) | 1.000 | 1.225 | *** | 0.39 |

*Note: Sig. refers to the difference between platforms.*

Table 7: Facebook vs. Knesset Data (2018-2021)

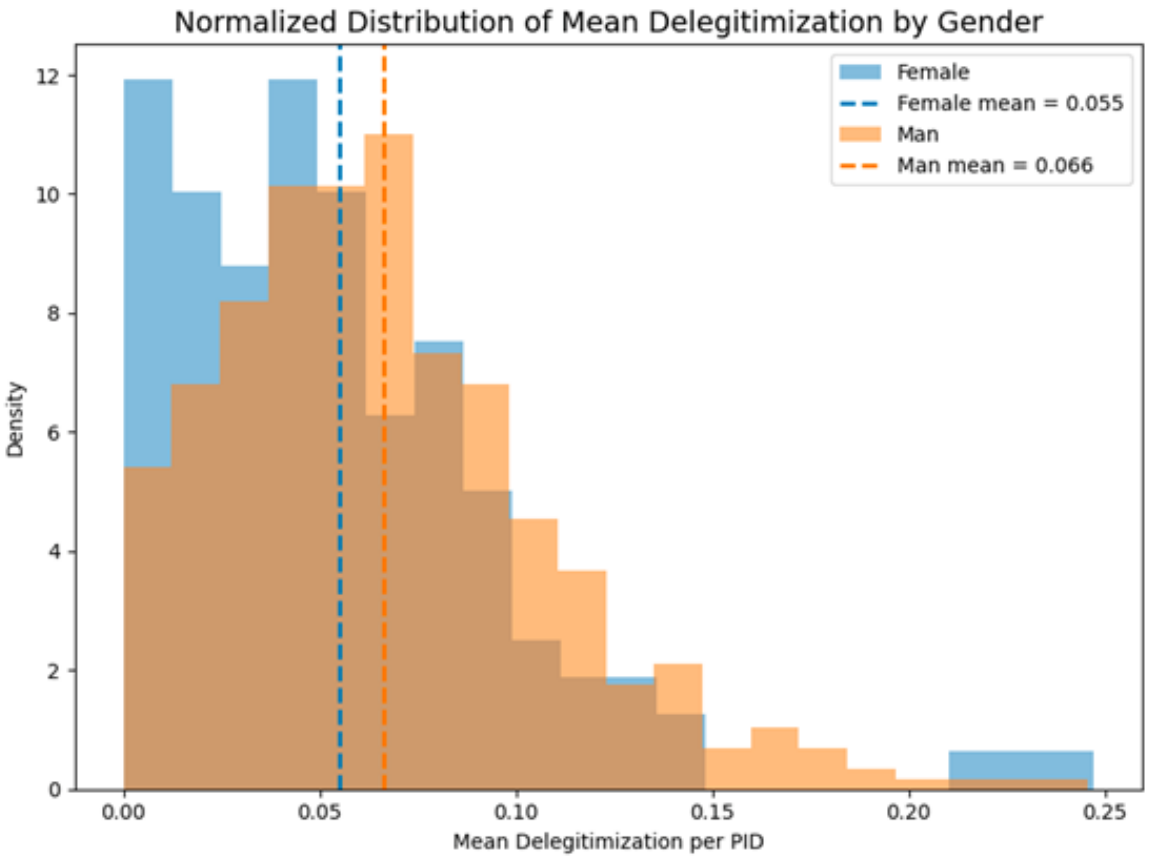

Figure 3: Knesset PDD by Gender

the Knesset dataset. Figure 3 presents the normalized density of PDD expression among male (orange) and female (blue) politicians. The results indicate that male politicians exhibit slightly higher average levels of PDD (mean = 0.066) than their female counterparts (mean = 0.055). Although the difference in means is modest, it is statistically significant ($t = 2.613$, $p = 0.010$, $|d| = 0.268$).[4]

To explore political variation in the use of PDD (RQ4), we compare the distribution of mean PDD scores across political blocs (right, center, and left) in both Knesset speeches and Facebook posts. Figure 4 presents three panels: the full Knesset corpus from 1992-2023 (left), the Facebook dataset from 2018-2021 (center), and a subset of the Knesset data for the same 2018-2021 period (right).

In the long-term Knesset data, right-wing politicians display the highest average level of delegitimization (mean = 0.074), followed by the left (0.062) and center (0.048). However, when examining the 2018-2021 period, a different picture emerges: left-leaning politicians exhibit the highest average PDD score in both platforms (Facebook mean = 0.080, Knesset mean = 0.089), followed by the center and right-wing politicians. These findings suggest that while the right dominates in PDD over the long term, left-wing actors may become more rhetorically aggressive in specific periods. The variation across platforms and time highlights the importance of contextual factors in shaping the rhetorical strategies of different blocs.

To dive deeper into the distinction between political blocs, we examine the targets used within PDD in the Knesset dataset. Using the weighted log odds method proposed by Monroe et al. (2008), we identify the PDD target tokens most strongly associated with each political bloc. Table 8 presents the top distinctive tokens per bloc, with translations.

Across all blocs, PDD is primarily directed at individual political figures. Prominent names such as "Netanyahu," "Ben Gvir," and "Lapid" appear frequently, indicating that personal attacks are a common rhetorical strategy. However, what stands out most is the unique pattern observed in the right-wing bloc: in addition to targeting individuals, right-leaning politicians disproportionately use group-level and party-related terms such as "the Left," "Meretz," and "Kadima." This suggests that the right's delegitimization strategy extends beyond individuals to symbolic and collective targets, reflecting a broader ideological framing. Thus, while PDD is common across blocs, only the right consistently frames entire political groups as illegitimate.

[4]Facebook data shows a similar pattern, see Appendix D.

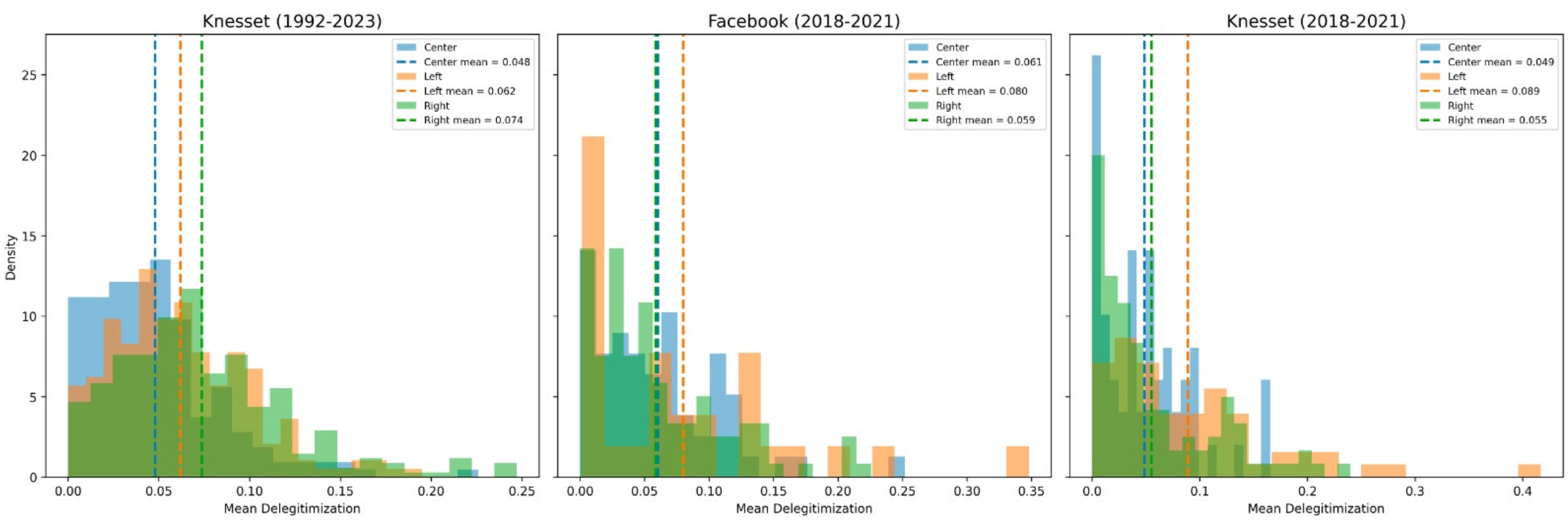

Figure 4: PDD Differences by Political Bloc, Knesset and Facebook Datasets

Finally, to assess how political context affects the use of PDD (RQ5), we begin by examining the weekly mean share of PDD within our Facebook dataset (Figure 5). This dataset allows for greater temporal granularity and captures short-term rhetorical shifts around key political events. The timeline includes four national elections in Israel, marked by vertical dashed lines. The data reveal clear spikes in PDD surrounding each election, suggesting that delegitimizing rhetoric intensifies during campaign periods - likely reflecting its strategic value for mobilizing supporters, discrediting opponents, and framing political conflict. Following each election, particularly after the formation of the unity government in April 2020, the volume of PDD drops markedly and remains relatively low throughout the governance period.

To further unpack these dynamics, we compare mean PDD levels for coalition and opposition actors before and after the government agreement (Figure 6). This analysis shows that while both groups reduce their use of delegitimizing rhetoric post-election, the drop is more substantial among coalition members (from 0.08 to 0.04) than among opposition members (from 0.11 to 0.08). Together, these findings highlight the contextual and strategic nature of PDD: it peaks during competitive election periods and recedes under conditions that prioritize coalition-building and institutional stability. Moreover, opposition actors consistently engage in more PDD than coalition members, indicating that their role outside government may incentivize more confrontational rhetoric.

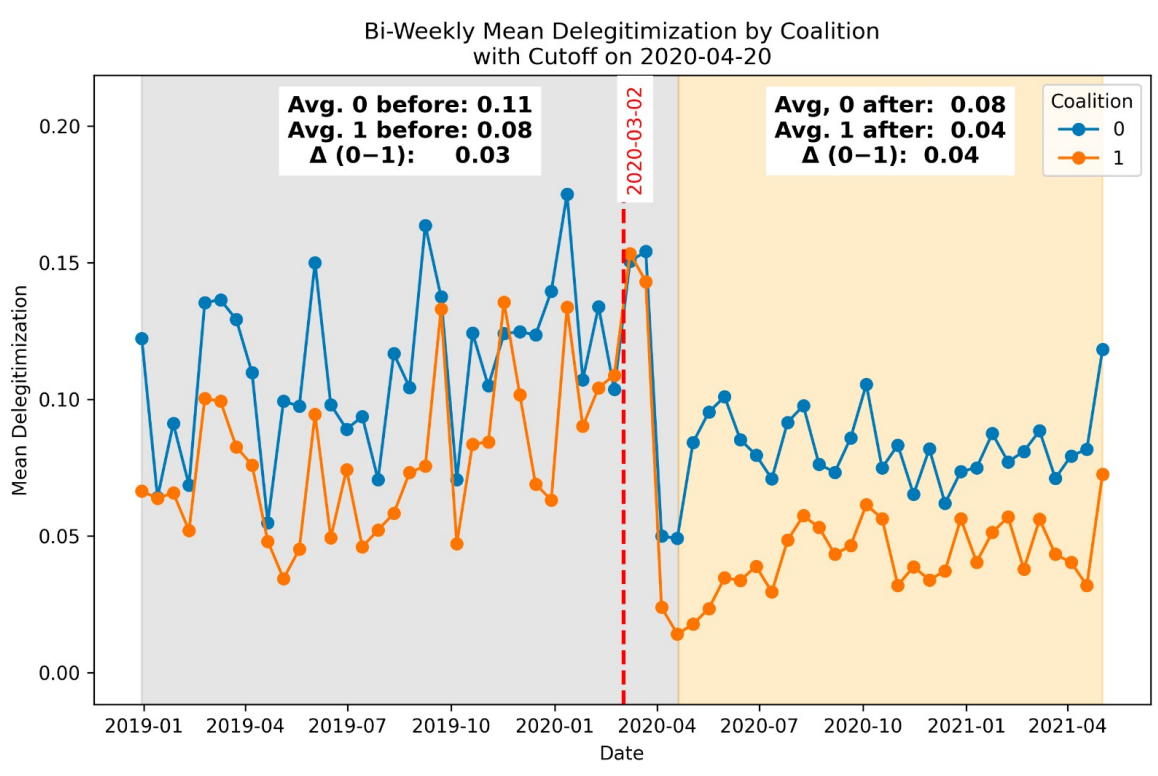

Figure 6: PDD Before and After 2020 Gov (FB)

## 8 Summary

In this paper, we introduce Political Delegitimization Discourse (PDD) as a novel framework for understanding how political actors employ symbolic attacks to undermine the legitimacy of opponents. We curate and manually annotate a large-scale, Hebrew-language corpus of 10,410 sentences drawn from Knesset speeches, Facebook posts, and leading news outlets, identifying 1,812 instances of PDD and richly annotating 642 of them for intensity, incivility, target type, and affective framing. We propose a two-stage detection pipeline that combines finetuned encoder models

| Top Delegitimized Targets | | |
|---|---|---|
| **By Speakers from the Left** | **By Speakers from the Center** | **By Speakers from the Right** |
| (Netanyahu) [P] | (Ben Gvir) [P] | (The Left) |
| (Sharon) [P] | (Amsalem) [P] | (Meretz) |
| (The government) | (Netanyahu) [P] | (Rabin) [P] |
| (Rivlin) [P] | (Smotrich) [P] | (Bennet) [P] |
| (Gideon Ezra) [P] | (Blue and White) | (Lapid) [P] |
| (Landau) [P] | (Rothman) [P] | (Peres) [P] |
| (Limor Livnat) [P] | (Gafni) [P] | (Sarid) [P] |
| (Ze'ev) [P] | (Karhi) [P] | (Beilin) [P] |
| (Eitan) [P] | (Akunis) [P] | (Mansour Abbas) [P] |
| (Kleiner) [P] | (Litzman) [P] | (Barak) [P] |
| (Silvan Shalom) [P] | (Shas) | (Kadima) |

*Note: Terms are ranked from most to least distinguishing for each bloc. [P] indicates the target is a person.*

Table 8: Top Distinguishing Terms Used as Targets in PDD by Speakers from Each Political Bloc (Knesset Dataset)

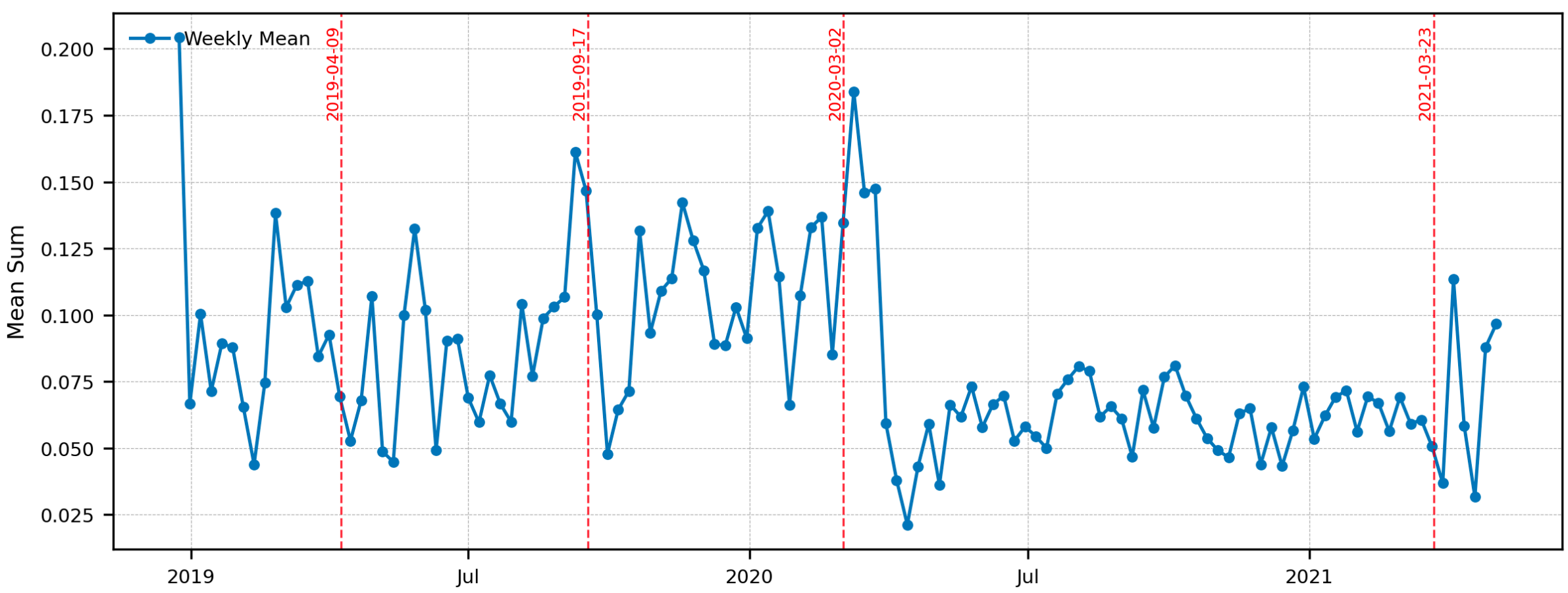

Figure 5: Share of Delegitimization in Facebook Posts

(e.g., DictaBERT) and decoder LLMs (DictaLM 2.0), achieving up to 0.74 $F_1$ on binary PDD detection and 0.67 macro-$F_1$ on multi-attribute classification. Applying our best model to longitudinal and cross-platform data, we document a clear upward trend in PDD over three decades, higher prevalence and intensity on social media compared to parliamentary debate, gender-based differences (with male politicians using PDD more frequently), and ideological patterns that vary over time and context-spiking during election campaigns and receding during coalition periods.

Our work extends beyond existing NLP tasks such as toxicity detection, stance detection, and targeted sentiment analysis. Whereas these approaches typically rely on surface-level cues of hostility or polarity, PDD captures a distinct, intent-based phenomenon: the denial of political legitimacy. By operationalizing this concept, we contribute (1) a novel task definition grounded in political theory, (2) a new annotated Hebrew-language corpus of political texts, and (3) a dedicated classification pipeline. Together, these contributions expand the NLP community's ability to detect subtle forms of harmful political speech that existing models systematically overlook.

Furthermore, our framework has broader applicability. Methodologically, our annotation scheme and modeling pipeline can be adapted to other contexts of democratic erosion to enable comparative analyses of how PDD manifests across regimes. Substantively, our approach provides new tools for studying how elite discourse interacts with phenomena such as affective polarization, identity threat, and identity deconstruction, particularly in settings where political actors challenge the legitimacy of opposition groups, institutions, or electoral outcomes. By bridging computational methods with political theory, we demonstrate both the feasibility of automated PDD analysis and its value for tracking democratic discourse dynamics at scale.

## Limitations

A key limitation of our study concerns its scope and transferability. While our Hebrew-language corpus and Israeli political focus enable an in-depth case study, the findings may not directly generalize to other linguistic or institutional and political contexts. Likewise, our social-media analysis is limited to Facebook posts from 2018-2021, leaving unexamined discourse on other platforms and in more recent periods.

The transfer of our approach to other languages and political contexts may pose several challenges. What constitutes delegitimization is deeply shaped by local political dynamics and social cleavages. For example, references that may be highly delegitimizing in the Israeli context (e.g., mentions of settlers or Arab citizens) may not carry the same meaning elsewhere. Adapting our framework thus requires linguistic expertise and domain knowledge to identify context-specific delegitimizing cues. Thus, while our general annotation scheme and the conceptualization of the PDD task is applicable to other contexts, the specific annotated instances may not be relevant to other political environments.

Another limitation stems from our annotation approach. We rely on sentence-level annotation without incorporating broader discourse context, which may miss cases where delegitimization emerges only across multiple sentences or conversational turns. Subtle rhetorical strategies - such as irony, implication, or metaphor, may remain difficult to capture at the sentence level.

Moreover, while disagreements between the human coders were resolved during the first phase of annotation, the second-stage annotation of fine-grained characteristics was conducted by a single annotator. Although this annotator is a senior domain expert in political delegitimization discourse, the use of a single annotator in this stage could introduce bias.

Finally, our analysis is descriptive and correlational: we document how PDD varies over time, by gender, and across blocs, but do not establish causal pathways. Future work should integrate richer contextual signals, extend to diverse data sources, and examine how delegitimizing rhetoric influences audience attitudes.

## Acknowledgments

This study was funded by Israel Science Foundation grant (No. 2315/18) for the Center of Excellence "Looking Beyond the Crisis of Democracy: Patterns of Representation in Israeli Elections." We thank Michal Shamir, Shaul R. Shenhav, Yael R. Kaplan, and all members of the DeepStory lab at the Hebrew University of Jerusalem for insightful advice and comments.

## References

Abeer ALDayel and Walid Magdy. 2021. Stance detection on social media: State of the art and trends. *Information Processing & Management*, 58(4):102597.

Benedetta Baldi and Ludovico Franco. 2015. (de) legitimization strategies in the "austere prose" of palmiro togliatti. *Quaderni di Linguistica e Studi Orientali*, 1:139–158.

Benedetta Baldi, Ludovico Franco, and Leonardo M Savoia. 2019. Alternative truths and delegitimization pragmatic strategies around the 2018 italian elections. *Journal of Language Aggression and Conflict*, 7(2):293–320.

Daniel Bar-Tal. 1989. Delegitimization: The extreme case of stereotyping and prejudice. In D. Bar-Tal, C. Graumann, A. W. Kruglanski, and W. Stroebe, editors, *Stereotyping and Prejudice*, pages 169–182. Springer Verlag, New York.

Daniel Bar-Tal. 1990. Causes and consequences of delegitimization: Models of conflict and ethnocentrism. *Journal of Social issues*, 46(1):65–81.

Jamie Bartlett. 2014. Populism, social media and democratic strain. *European populism and winning the immigration debate*, pages 99–116.

Wojciech Baryla, Bogdan Wojciszke, and Aleksandra Cichocka. 2015. Legitimization and delegitimization of social hierarchy. *Social Psychological and Personality Science*, 6(6):669–676.

María Berrocal. 2019. Delegitimization strategies in czech parliamentary discourse. In I. Laitinen and H. D. T. T. Nguyen, editors, *Political Discourse in Central, Eastern and Balkan Europe*, pages 119–146. John Benjamins Publishing Company.

Lawrence Buell. 1998. Toxic discourse. *Critical inquiry*, 24(3):639–665.

Kateryna Burovova and Mariana Romanyshyn. 2024. Computational analysis of dehumanization of ukrainians on russian social media. In *Proceedings of the 8th Joint SIGHUM Workshop on Computational Linguistics for Cultural Heritage, Social Sciences, Humanities and Literature (LaTeCH-CLfL 2024)*, pages 28–39.

Ernesto Calvo, Tomás Ventura, Natalia Aruguete, and Silvio Waisbord. 2023. Winning! election returns and engagement in social media. *PLOS ONE*, 18(3):e0281475.

Piotr Cap. 2008. Towards the proximization model of the analysis of legitimization in political discourse. *Journal of Pragmatics*, 40(1):17–41.

Paul Chilton. 2004. *Analysing Political Discourse: Theory and Practice*. Routledge, London.

Giovanni Da San Martino, Alberto Barrón-Cedeño, Henning Wachsmuth, Rostislav Petrov, and Preslav Nakov. 2020. SemEval-2020 task 11: Detection of propaganda techniques in news articles. In *Proceedings of the Fourteenth Workshop on Semantic Evaluation*, pages 1377–1414, Barcelona (online). International Committee for Computational Linguistics.

Pieter Delobelle, Murilo Cunha, Eric Massip Cano, Jeroen Peperkamp, and Bettina Berendt. 2019. Computational ad hominem detection. In *Proceedings of the 57th Annual Meeting of the Association for Computational Linguistics: Student Research Workshop*, pages 203–209.

Tim Dettmers, Artidoro Pagnoni, Ari Holtzman, and Luke Zettlemoyer. 2023. Qlora: Efficient finetuning of quantized llms. *Preprint*, arXiv:2305.14314.

Jacob Devlin, Ming-Wei Chang, Kenton Lee, and Kristina Toutanova. 2019. Bert: Pre-training of deep bidirectional transformers for language understanding. In *Proceedings of the 2019 Conference of the North American Chapter of the Association for Computational Linguistics: Human Language Technologies*, pages 4171–4186.

J. L. Egelhofer, L. Aaldering, and S. Lecheler. 2021. Delegitimizing the media? analyzing politicians' media criticism on social media. *Journal of Language and Politics*, 20(5):653–675.

Mattias Ekman and Andreas Widholm. 2015. Politicians as media producers: Current trajectories in the relation between journalists and politicians in the age of social media. *Journalism practice*, 9(1):78–91.

Lukas Erhard, Sara Hanke, Uwe Remer, Agnieszka Falenska, and Raphael Heiko Heiberger. 2025. Popbert. detecting populism and its host ideologies in the german bundestag. *Political Analysis*, 33(1):1–17.

Jeremy A. Frimer, Harleen Aujla, Matthew Feinberg, Linda J. Skitka, Karl Aquino, Johannes C. Eichstaedt, and Robb Willer. 2023. Incivility is rising among american politicians on twitter. *Social Psychological and Personality Science*, 14(2):259–269.

Savitri Gadavanij. 2020. Contentious polities and political polarization in thailand: Post-thaksin reflections. *Discourse & Society*, 31(1):44–63.

Gemma Team. 2024. Gemma 2: Improving open language models at a practical size. *Preprint*, arXiv:2408.00118.

Bryan T Gervais, Connor Dye, and Amber Chin. 2025. Incivility or invalidity? evaluating perspective api scores as a measure of political incivility. *American Politics Research*.

Noam Gidron, James Adams, and Will Horne. 2020. *American Affective Polarization in Comparative Perspective*. Cambridge University Press, Cambridge.

Donald P. Green, Bradley Palmquist, and Eric Schickler. 2004. *Partisan Hearts and Minds: Political Parties and the Social Identities of Voters*. Yale University Press, New Haven.

Ivan Habernal, Henning Wachsmuth, Iryna Gurevych, and Benno Stein. 2018. Before name-calling: Dynamics and triggers of ad hominem fallacies in web argumentation. *arXiv preprint arXiv:1802.06613*.

Ryan W. Hansen. 2023. You've never been welcome here: Exploring the relationship between exclusivity and incivility in online forums. *Journal of Information Technology & Politics*, 20(2):139–153.

Martin Haselmayer, Sarah C Dingler, and Marcelo Jenny. 2022. How women shape negativity in parliamentary speeches—a sentiment analysis of debates in the austrian parliament. *Parliamentary Affairs*, 75(4):867–886.

Jack Holland and Kathryn A. M. Wright. 2017. The double delegitimisation of julia gillard: Gender, the media, and australian political culture. *Australian Journal of Politics & History*, 63(4):588–602.

Shanto Iyengar and Masha Krupenkin. 2018. The strengthening of partisan affect. *Political Psychology*, 39:201–218.

Shanto Iyengar, Yphtach Lelkes, Matthew Levendusky, Neil Malhotra, and Sean J. Westwood. 2019. The origins and consequences of affective polarization in the united states. *Annual Review of Political Science*, 22(1):129–146.

Md Saroar Jahan and Mourad Oussalah. 2023. A systematic review of hate speech automatic detection using natural language processing. *Neurocomputing*, 546:126232.

Christopher F. Karpowitz and Tali Mendelberg. 2014. *The Silent Sex: Gender, Deliberation, and Institutions*. Princeton University Press, Princeton.

Herbert C Kelman. 2001. Reflections on social and psychological processes of legitimization and delegitimization. *The psychology of legitimacy: Emerging perspectives on ideology, justice, and intergroup relations*, pages 54–73.

Christopher Klamm, Ines Rehbein, and Simone Paolo Ponzetto. 2023. Our kind of people? detecting populist references in political debates. In *Findings of the association for computational linguistics: EACL 2023*, pages 1227–1243.

Steven Levitsky and Daniel Ziblatt. 2019. *How democracies die*. Crown.

Tsung-Yi Lin, Priya Goyal, Ross Girshick, Kaiming He, and Piotr Dollár. 2017. Focal loss for dense object detection. In *Proceedings of the IEEE International Conference on Computer Vision (ICCV)*, pages 2980–2988.

Lilliana Mason. 2018. *Uncivil Agreement: How Politics Became Our Identity*. University of Chicago Press, Chicago.

Julia Mendelsohn, Yulia Tsvetkov, and Dan Jurafsky. 2020. A framework for the computational linguistic analysis of dehumanization. *Frontiers in artificial intelligence*, 3:55.

Saif M. Mohammad, Parinaz Sobhani, and Svetlana Kiritchenko. 2017. Stance and sentiment in tweets. *ACM Trans. Internet Technol.*, 17(3).

Burt L Monroe, Michael P Colaresi, and Kevin M Quinn. 2008. Fightin'words: Lexical feature selection and evaluation for identifying the content of political conflict. *Political Analysis*, 16(4):372–403.

Guy Mor-Lan, Effi Levi, Tamir Sheafer, and Shaul R. Shenhav. 2024. IsraParlTweet: The israeli parliamentary and Twitter resource. In *Proceedings of the 2024 Joint International Conference on Computational Linguistics, Language Resources and Evaluation (LREC-COLING 2024)*, pages 9372–9381, Torino, Italia. ELRA and ICCL.

John T Nockleby. 1994. Hate speech in context: The case of verbal threats. *Buff. L. Rev.*, 42:653.

John H Parmelee and Shannon L Bichard. 2011. *Politics and the Twitter revolution: How tweets influence the relationship between political leaders and the public*. Lexington books.

Qwen Team. 2025. Qwen3.

Antonio Reyes. 2011. Strategies of legitimization in political discourse: From words to actions. *Discourse & Society*, 22(6):781–807.

Ludovic Rheault, Erica Rayment, and Andreea Musulan. 2019. Politicians in the line of fire: Incivility and the treatment of women on social media. *Research & Politics*, 6(1):2053168018816228.

Khalil Rinnawi. 2007. De-legitimization of media mechanisms: Israeli press coverage of the al aqsa intifada. *International Communication Gazette*, 69(2):149–178.

Naama Rivlin-Angert. 2025. *The Deconstruction of Political Identity: The Case of the Repudiating Left in Israel*. Ph.d. dissertation, Tel Aviv

University, School of Political Science, Government and International Affairs, Tel Aviv.

Andrew S Ross and Damian J Rivers. 2017. Digital cultures of political participation: Internet memes and the discursive delegitimization of the 2016 us presidential candidates. *Discourse, Context & Media*, 16:1–11.

Patrícia Rossini. 2022. Beyond incivility: Understanding patterns of uncivil and intolerant discourse in online political talk. *Communication Research*, 49(3):399–425.

Hind Saleh, Areej Alhothali, and Kawthar Moria. 2023. Detection of hate speech using bert and hate speech word embedding with deep model. *Applied Artificial Intelligence*, 37(1):2166719.

Francesco Screti. 2013. Defending joy against the popular revolution: legitimation and delegitimation through songs. *Critical Discourse Studies*, 10(2):205–222.

Amit Seker, Elron Bandel, Dan Bareket, Idan Brusilovsky, Refael Greenfeld, and Reut Tsarfaty. 2022. Alephbert: Language model pre-training and evaluation from sub-word to sentence level. In *Proceedings of the 60th Annual Meeting of the Association for Computational Linguistics (Volume 1: Long Papers)*, pages 46–56.

Vitaly Shalumov and Harel Haskey. 2023. Hero: Roberta and longformer hebrew language models. *Preprint*, arXiv:2304.11077.

Shaltiel Shmidman, Avi Shmidman, Amir DN Cohen, and Moshe Koppel. 2024. Adapting llms to hebrew: Unveiling dictalm 2.0 with enhanced vocabulary and instruction capabilities. *Preprint*, arXiv:2407.07080.

Shaltiel Shmidman, Avi Shmidman, and Moshe Koppel. 2023. Dictabert: A state-of-the-art bert suite for modern hebrew. *Preprint*, arXiv:2308.16687.

Rasmus Skytte. 2021. Dimensions of elite partisan polarization: Disentangling the effects of incivility and issue polarization. *British Journal of Political Science*, 51(4):1457–1475.

Mark C. Suchman. 1995. Managing legitimacy: Strategic and institutional approaches. *Academy of Management Review*, 20(3):571–610.

Milan W Svolik. 2019. Polarization versus democracy. *Journal of democracy*, 30(3):20–32.

Yuzhou Tao, Zhiqin Zhan, Han Zhou, Jingshi Kang, and Shaojing Sun and. 2024. Measuring chinese online populist discourse: an automated semantic text analysis method. *Chinese Journal of Communication*, 0(0):1–21.

Yannis Theocharis, Pablo Barberá, Zoltán Fazekas, and Sebastian Adrian Popa. 2020. The dynamics of political incivility on twitter. *Sage Open*, 10(2).

Cristian Tileagă. 2007. Ideologies of moral exclusion: A critical discursive reframing of depersonalization, delegitimization and dehumanization. *British journal of social psychology*, 46(4):717–737.

Teun A van Dijk. 1997. What is political discourse analysis. *Belgian journal of linguistics*, 11(1):11–52.

Teun A van Dijk. 1998. *Ideology: A multidisciplinary approach*. SAGE Publications Ltd, London, UK.

Erika van Elsas and Toine Fiselier. 2023. Conflict or choice? the differential effects of elite incivility and ideological polarization on political support. *Acta Politica*.

Theo van Leeuwen. 2007. Legitimation in discourse and communication. *Discourse & Communication*, 1(1):91–112.

Chiara Volpato, Francesca Durante, Alessandro Gabbiadini, Laura Andrighetto, and Silvia Mari. 2010. Picturing the other: Targets of delegitimization across time. *International Journal of Conflict and Violence*, 4(2):269–287.

Max Weber. 1958. The three types of legitimate rule. *Berkeley Publications in Society and Institutions*, 4(1):1–11.

Elke Winter. 2016. ‘immigrants don’t ask for self-government’: How multiculturalism is (de) legitimized in multinational societies. In *Migration and Divided Societies*, pages 45–62. Routledge.

Meishan Zhang, Yue Zhang, and Duy-Tin Vo. 2016. Gated neural networks for targeted sentiment analysis. In *Proceedings of the AAAI conference on artificial intelligence*, volume 30.

Karen Zhou, Alexander A Meitus, Milo Chase, Grace Wang, Anne Mykland, William Howell, and Chenhao Tan. 2024. Quantifying the uniqueness of donald trump in presidential discourse. *arXiv preprint arXiv:2401.01405*.

## A Full Annotation Guidelines

This Appendix presents the full annotation guidelines for human coders, including examples and detailed criteria. The human-annotated process was according to the following coding book:

A sentence will be recognized as PDD if the discourse is directed towards political groups (left or right), political actors (politicians or parties), or political institutions [hereinafter - the object] and based on an attack on the emotional identification with these groups and/or actors. The discourse is intended to produce negative feelings towards the object when there are no remarks at all to policy or rational arguments (any "persuasive" discourse that seeks to change a position or any ideological arguments do not fall within the definition of delegitimization). This type of discourse is based on an extremely negative characterization of the identity groups according to the following parameters:

- Expressing feelings of disgust and hatred towards the group, ridicule (e.g., "smollanim"), swearing words, or profanity.
- Arguments that challenge the group's right to exist, designed to muzzle them.
- Characterizing the group as deliberately harming the common good, in a way that constitutes a danger to society / the state.
- Denial of humanity, demonization of group members.
- Comparison and connection to other groups with a negative connotation (e.g., Nazis, Fascists, Arabs, Hilltop Youth).

## B Prompt Templates

The general prompt format used for fine-tuning the decoder-only models on all tasks is as follows:

```
{sentence}\n### Answer: {output}
```

Where the content and structure of output differs between the tasks:

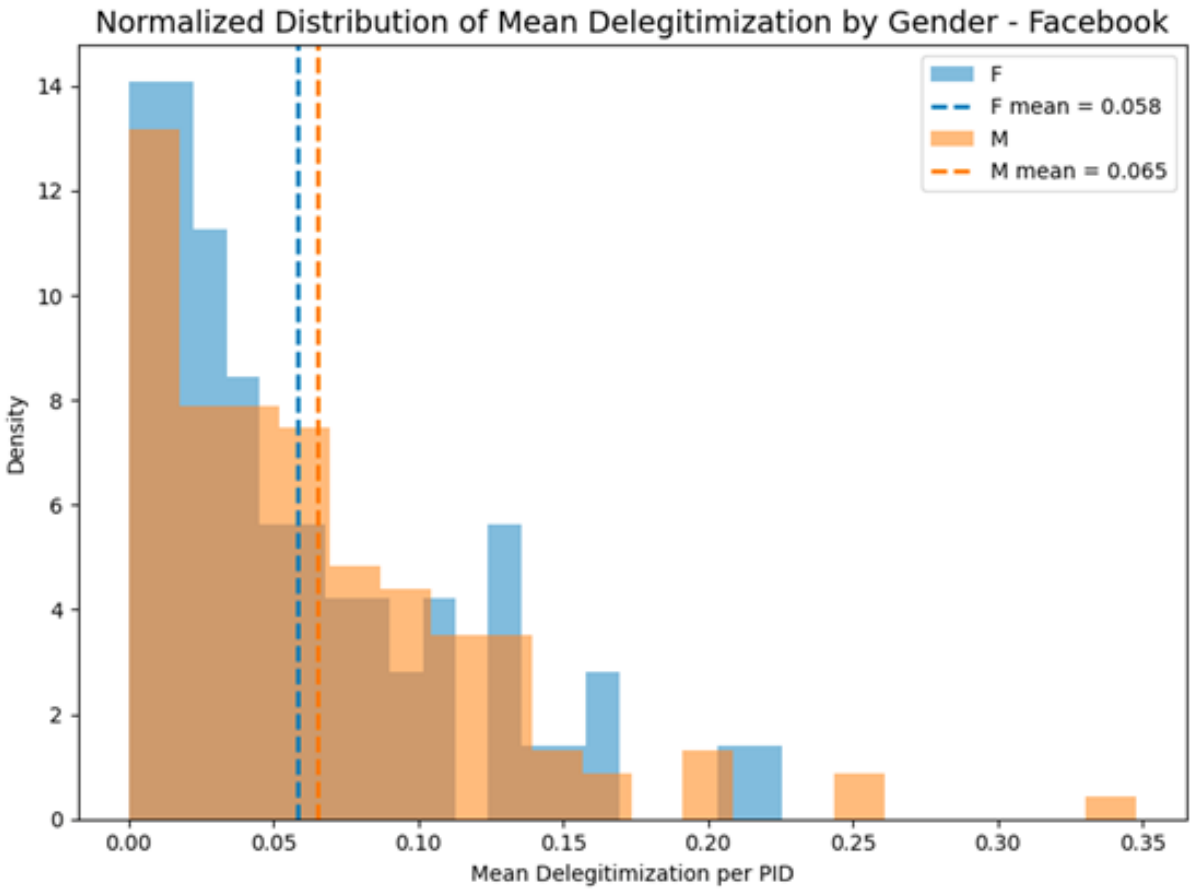

Figure 7: Normalized Distribution of Mean Delegitimization in Facebook Posts by Gender (2018–2021)

- In the binary PDD detection task, output is either 'Yes' or 'No' in Hebrew (/).
- In the PDD attribute classification task, output is a JSON dictionary whose keys are the target variable names (intensity, incivility, group, person, outgroup, common_good, institute) and whose values are 0/1 or 0/1/2 (for intensity).
- In the target span detection task, output is the input sentence with %%% tags added to mark target spans.

## C Gender Differences in PDD on Facebook

To validate the gender-based findings observed in the Knesset dataset, we replicate the analysis using the Facebook dataset covering the period 2018–2021. As in the main analysis, we compute the mean share of PDD sentences per speaker and compare the distributions by gender.

The results are consistent with those observed in parliamentary speech data. Figure 7 shows the normalized density of mean PDD per politician, disaggregated by gender. Male politicians exhibit a slightly higher average level of PDD (mean = 0.065) compared to female politicians (mean = 0.058). Although the absolute difference is small, it is significant, suggesting that male politicians tend to use delegitimizing rhetoric more frequently than their female counterparts, even in the unmediated and informal environment of social media.

## D Fine-tuning setup

All model fine-tuning is performed on an 80GB A100 Nvidia GPU, using huggingface transformers.

For decoder fine-tuning, the separator "### Answer:" is used to separate the input sentences from the output. The labels of the input and separator are loss-masked.

All experiments utilize AdamW optimizer with a linear scheduler. Default values of hyper-paramters are used everywhere except for learning rate (for encoders and for decoder LLMs) and loss type (for encoder models).

Decoder fine-tuning uses QLORA with 4bit quanitzation. LORA settings are rank of 256, and alpha value of 512. LORA layers are attached to all linear levels in the decoder models.

Each hyper-parameter configuration was trained once.

## E Data Release

The annotated data is released under cc-by-4.0 license. The data is publicly available on github at https://github.com/guymorlan/pdd/.

## F Annotation

The data has been annotated by the lead author and two academic colleagues. The authors have not received direct compensation. All annotators are Hebrew-speaking Israelis. Two of the annotators are women and one is a man.

## G Model Sizes

| Model | Parameters |
|---|---:|
| mBERT | 110 M |
| AlephBERT | 110 M |
| HeRo | 125 M |
| DictaBERT–base | 184 M |
| DictaBERT–large | 340 M |
| Gemma–2B | 2 B |
| Gemma–9B | 9 B |
| Qwen3–8B | 8.2 B |
| DictaLM 2.0 | 7 B |

Table 9: Model sizes (number of parameters) for all models used in this paper.

## H Preprocessing

Sentence segmentation was performed using the Stanza package.